\documentclass[letterpaper]{article} 
\usepackage[preprint]{aaai2027}
\usepackage[hyphens]{url}  
\usepackage{graphicx} 
\usepackage{natbib}  
\usepackage{caption} 
\usepackage{algorithm}
\usepackage{algorithmic}

\usepackage{newfloat}
\usepackage{listings}
\DeclareCaptionStyle{ruled}{labelfont=normalfont,labelsep=colon,strut=off} 
\floatstyle{ruled}
\newfloat{listing}{tb}{lst}{}
\floatname{listing}{Listing}

\usepackage{booktabs}

\usepackage{amsmath}
\usepackage{xspace}
\usepackage{amssymb}
\usepackage{mathtools}
\usepackage{amsthm}
\usepackage{url} 
\usepackage{enumitem}
\usepackage{arydshln}
\usepackage{booktabs}  
\usepackage{colortbl}  
\usepackage{graphicx}   
\usepackage{xcolor}  
\definecolor{mygray}{gray}{0.9}  
\usepackage{algorithm}
\usepackage{listings}
\usepackage{graphicx}
\usepackage{makecell}
\usepackage{multirow}

\usepackage{booktabs}
\usepackage{adjustbox}
\usepackage{etoolbox}
\theoremstyle{plain}

\theoremstyle{definition}

\theoremstyle{remark}

\usepackage{xcolor}
\usepackage{booktabs}
\usepackage{graphicx} 
\definecolor{mygray}{gray}{0.5}
\usepackage{color}
\usepackage{colortbl}
\usepackage{array}
\usepackage[table]{xcolor}
\usepackage{pifont}
\usepackage[marginal]{footmisc}
\usepackage{booktabs}
\usepackage{multirow}
\usepackage{graphicx}
\newcommand{\methodname}{\textsc{TRAM}\xspace}
\title{TRAM: Enhancing Multimodal Reasoning with Trajectory-Derived \\
Auxiliary Memory}

\author{
    Kang Liu, Zijing Wang, Yongkang Liu, Mengjie Zhao, Xiaocui Yang, Shi Feng, Yifei Zhang, Daling Wang
}
\affiliations{
    Northeastern University, China\\
    lk\_stu\_neu@163.com
}

\begin{document}

\maketitle

\begin{abstract}
Multimodal Large Reasoning Models (MLRMs) have achieved strong performance on
tasks requiring visual understanding and multi-step inference. However, as
reasoning trajectories grow, models may become less effective at using
information established earlier in the context, increasing the risk of
reasoning errors. Existing approaches primarily address this problem by sustaining visual
grounding throughout reasoning. However, reasoning also transforms visual
observations into task-specific relations, constraints, and intermediate
conclusions whose influence may weaken over long trajectories. Our attribution
analysis suggests that correctness is not consistently separated by image
attribution alone, but is more closely associated with whether trajectories
retain and integrate such reasoning-derived information across stages. Motivated by this, we introduce
TRAM (TRajectory-derived Auxiliary Memory), a training-free method that
augments standard decoding with an auxiliary memory pathway derived from the
model's own reasoning trajectory. TRAM consolidates completed reasoning into a
compact latent memory, updates it online through fast and slow recurrent
streams, and feeds it back into selected decoder layers through a lightweight
residual pathway. Experiments across four MLRM variants on eight benchmarks show that TRAM
improves performance over vanilla decoding on mathematical, scientific, and
general visual reasoning tasks without additional training.
\end{abstract}



\section{Introduction}
Multimodal Large Reasoning Models (MLRMs) extend vision-language modeling from direct perception to deliberate visual reasoning.
By generating intermediate reasoning traces before the final answer, these models have achieved strong performance on tasks that require both visual understanding and multi-step inference
\citep{bai2025qwen3,wang2025internvl3,huang2025vision,meng2025mm}.
Nevertheless, as reasoning trajectories grow, models may fail to consistently use information established earlier in the context, causing later reasoning to deviate from an otherwise sound trajectory
\citep{liu2024lost,liu2026sinktrack}.

A common explanation in existing work is that the growing textual trajectory gradually weakens the contribution of the input image to later reasoning, making subsequent generation less grounded in visual evidence and more prone to errors.
Recent studies have therefore explored ways to sustain visual grounding throughout the reasoning process.
Some methods use reinforcement learning
\citep{yang2026look,huang2025spotlight,wang2026visually} or learnable visual
pathways to encourage continued reliance on the image
\citep{huang2026persistent}.
Inference-time methods further strengthen grounding through visually
conditioned decoding calibration and visual-representation steering
\citep{leng2024mitigating,li2025hidden,xu2026thinking}.
Other methods revisit the input image during reasoning
\citep{ghosal2026visref,gao2025interleaved,zhang2026see}.
Despite their different implementations, these approaches share the goal of strengthening the role of the original image throughout reasoning.

\begin{figure*}[t]
    \centering
    \includegraphics[width=0.98\textwidth]{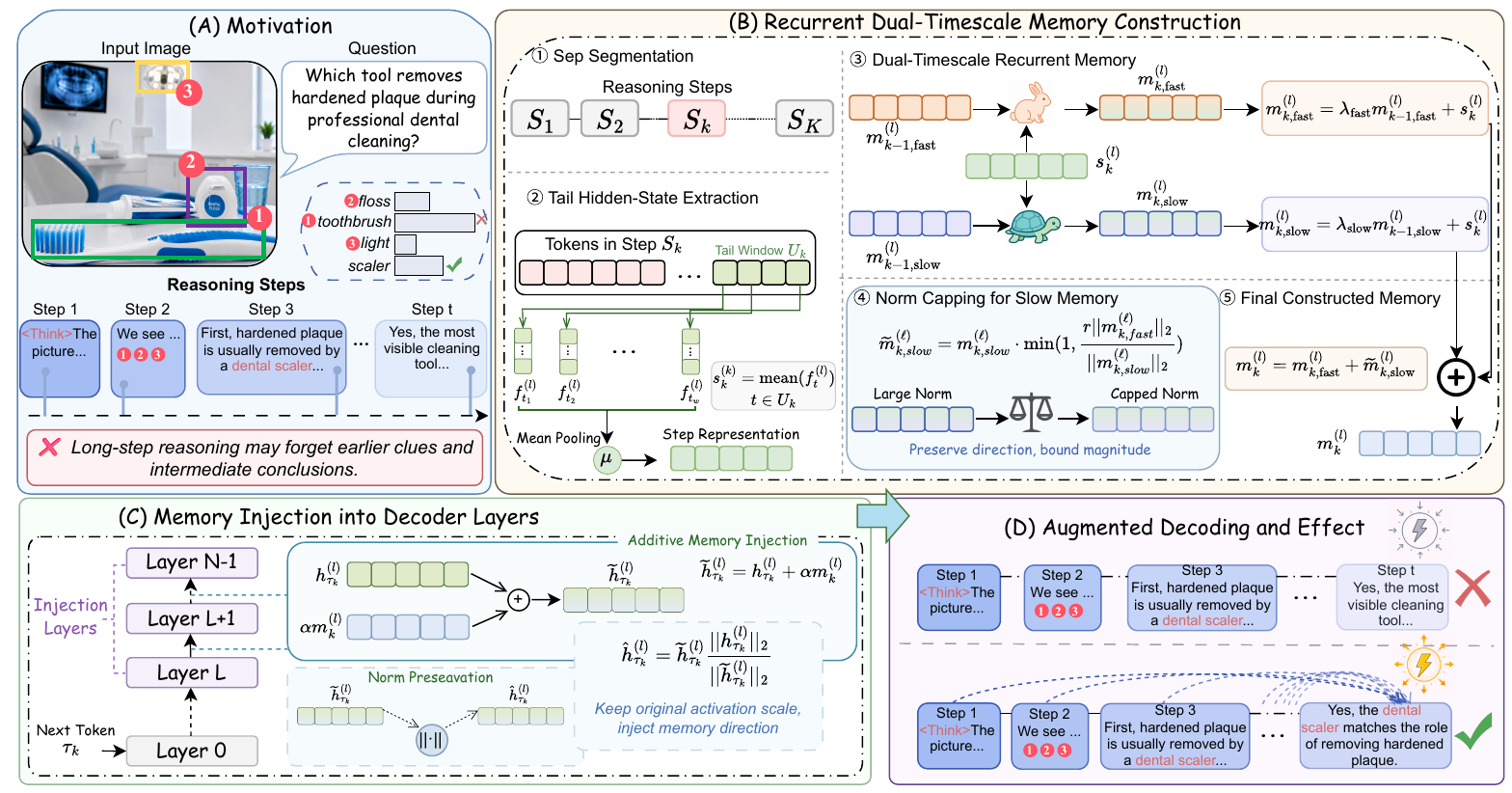}
    \caption{
    Overview of \methodname: (A) Motivation, (B) Recurrent Dual-Timescale
    Memory Construction, (C) Memory Injection into Decoder Layers, and (D)
    Augmented Decoding and Effect.
    }
    \label{fig:overview}
\end{figure*}

However, reasoning involves more than repeatedly consulting the original image.
As reasoning unfolds, the model transforms visual observations into
task-specific relations, constraints, and intermediate conclusions not
explicitly contained in the image. As illustrated in
Figure~\ref{fig:overview}(A), such information may be formed early, while its
influence may also weaken as the trajectory grows. Replaying or reinforcing
the image restores access to the original visual evidence, but does not carry
forward the reasoning information already derived from it. We further examine
how visual evidence and prior reasoning contribute across the trajectory
through attribution analysis in the next section. This analysis shows that, as
reasoning progresses, image attribution decreases in both correct and incorrect
trajectories, with no consistent difference between them. Instead, it reveals a
clearer difference: correct trajectories retain and integrate information formed
across multiple reasoning stages, whereas incorrect trajectories show more
limited use of such accumulated information.

Motivated by these findings, we introduce \textbf{TRAM} (\textbf{TR}ajectory-derived \textbf{A}uxiliary \textbf{M}emory),
a training-free method that augments standard decoding with a compact memory
constructed from the model's own reasoning trajectory. During reasoning, \methodname consolidates completed
reasoning into a compact latent memory state and updates it online as the
trajectory unfolds. The resulting memory is fed back into selected decoder
layers through a lightweight residual pathway, making information formed
earlier in the trajectory directly available to subsequent computation. The
memory is updated through two recurrent streams with different temporal
scales: the fast stream remains responsive to recent reasoning, whereas the
slow stream retains information over a longer horizon. This dual-timescale
update balances recent responsiveness with longer-range retention, helping
reasoning-derived information remain influential as the trajectory grows. The
intervention is sparse and minimally invasive, requiring no additional
training, parameter updates, external verifier, or modification to the
attention mechanism. Experiments across four MLRM variants on eight benchmarks show that \methodname
improves performance over vanilla decoding on mathematical, scientific, and
general visual reasoning tasks. Our contributions are threefold:
\begin{itemize}
    \item We conduct an attribution analysis of multimodal reasoning trajectories,
    revealing that image attribution alone does not consistently distinguish
    correct from incorrect trajectories, while correct trajectories show broader
    integration of information formed across multiple reasoning stages.

    \item We propose \textbf{\methodname}, a training-free method that augments
    standard decoding with a compact trajectory-derived memory, constructed from
    completed reasoning through fast and slow recurrent streams.

    \item We evaluate \methodname across four MLRM variants and eight reasoning
    benchmarks, showing higher mean accuracy than vanilla decoding without
    additional training.
\end{itemize}

\section{Reasoning Trajectory Analysis}
\label{sec:preliminary-analysis}

Long-form multimodal reasoning proceeds from visual evidence through
intermediate reasoning to a final answer. We trace attribution along this
process and compare correct and incorrect trajectories.

\begin{figure*}[t]
    \centering
    \includegraphics[width=0.242\textwidth,height=1.15in]{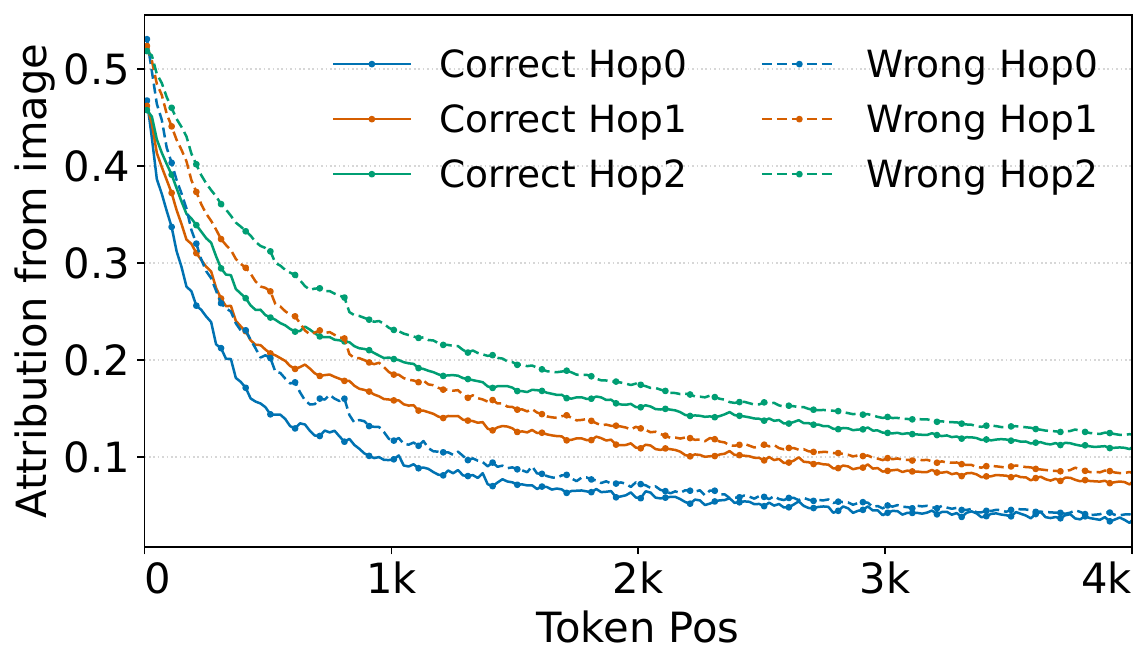}\hfill
    \includegraphics[width=0.242\textwidth,height=1.15in]{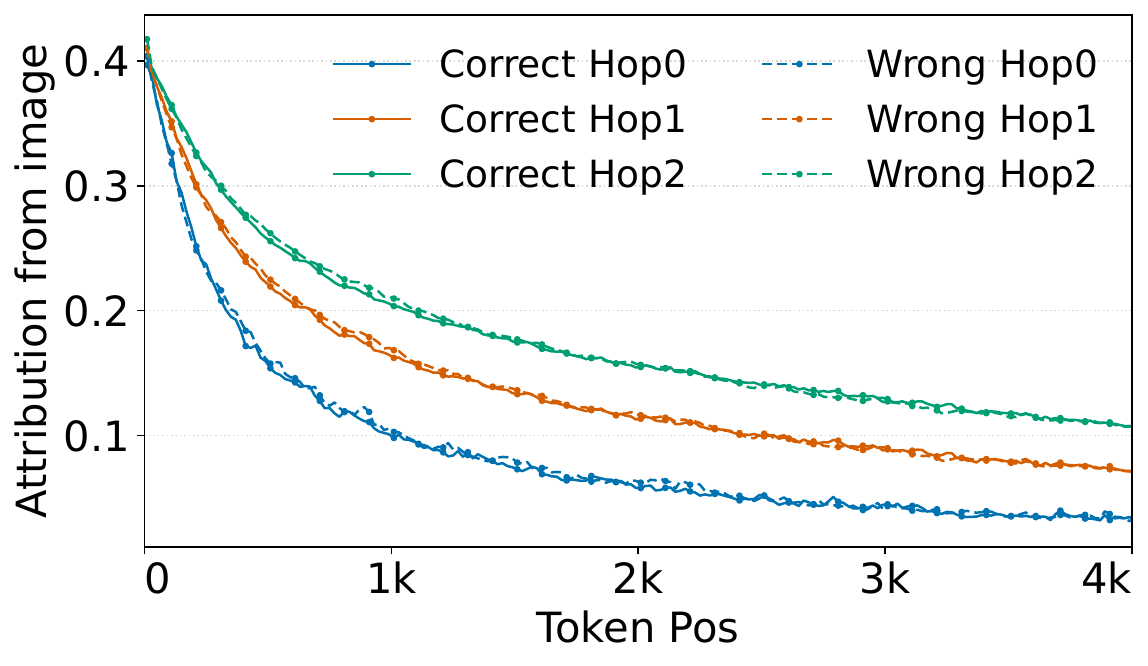}
    \hfill
    \includegraphics[width=0.242\textwidth,height=1.15in]{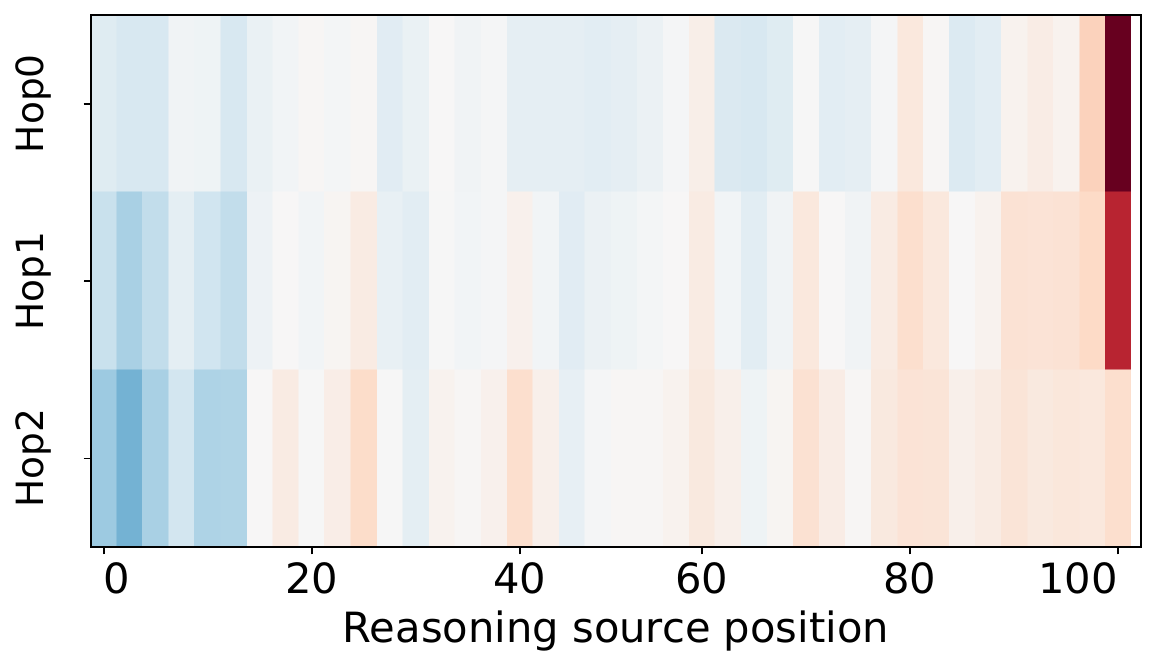}\hfill
    \includegraphics[width=0.242\textwidth,height=1.15in]{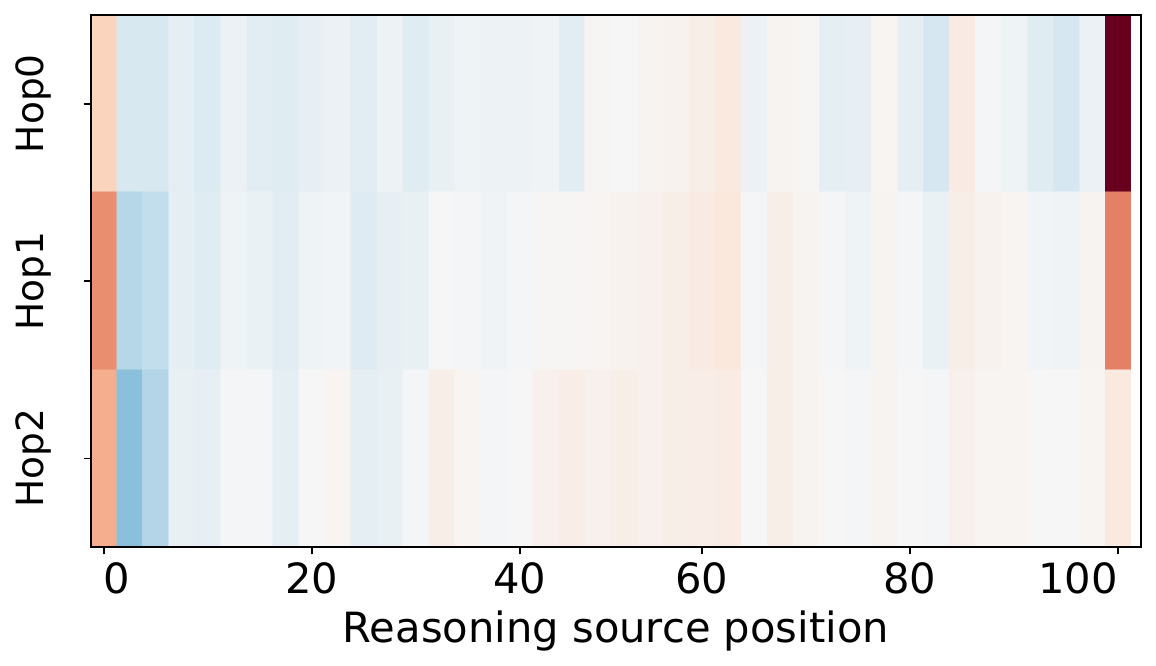}
    \caption{Attribution analysis of reasoning traces. (a) Image attribution
    across reasoning positions. (b) Correct-minus-incorrect answer attribution
    across relative source positions.}
    \label{fig:hop-motivation}
\end{figure*}

\paragraph{Attribution Setup.}
To estimate how earlier inputs and reasoning tokens contribute to a target, we
use FlashTrace \citep{pan2026towards}, since attention weights do not directly
quantify their contribution to the model output
\citep{jain2019attention}. FlashTrace recursively propagates attribution
through intermediate reasoning tokens to earlier source positions, including
visual tokens. We denote the $k$-hop attribution from a source span
$\mathcal{S}$ to a target span $\mathcal{T}$ as
$\operatorname{Hop}_k(\mathcal{S}\!\rightarrow\!\mathcal{T})$, where $k$ is
the recursion depth; larger $k$ captures more indirect dependencies. We analyze equal-sized sets of correct and incorrect vanilla traces generated by
Qwen3-VL-4B and InternVL3.5-4B on MathVision
\citep{wang2024measuring}, with further details provided in the Appendix. Two observations emerge:

\textbf{1) Image contribution persists but does not separate correctness.}
Figure~\ref{fig:hop-motivation}(a) shows that image attribution decreases as
reasoning progresses but remains measurable at later positions. Larger $k$
reveals additional visual influence mediated through intermediate reasoning
tokens. However, correct and incorrect traces show no consistent separation:
their curves largely overlap for InternVL3.5-4B, while incorrect traces can
exhibit higher attribution for Qwen3-VL-4B. Thus, weakened visual contribution
alone is insufficient to characterize incorrect reasoning.

\textbf{2) Correct and incorrect answers draw differently on the
reasoning history.}
Figure~\ref{fig:hop-motivation}(b) shows how the final answer draws on prior
reasoning. We divide each trajectory into 40 relative-position bins and
visualize normalized attribution differences between correct and incorrect
traces. At $k=0$, incorrect traces show higher attribution across much of the
trajectory, whereas correct traces concentrate near the final answer. As these
dependencies are recursively traced to earlier sources, attribution in
incorrect traces shifts toward the beginning and remains weak in the middle.
Their broad dependence at $k=0$ may thus reflect propagation of early
information rather than integration of distinct intermediate information.
Correct traces instead retain greater attribution at multiple middle-to-late
positions, indicating that relations, constraints, and intermediate
conclusions formed throughout reasoning contribute to the final answer.

\section{Method}

Based on the observations above, which show that correct trajectories better
integrate information formed across multiple reasoning stages, we introduce
\methodname, a training-free test-time method that augments standard decoding
with a compact memory constructed from the model's own reasoning trajectory.
As illustrated in Figure~\ref{fig:overview}, \methodname consists of two
components: (1) \textbf{Recurrent Dual-Timescale Memory}, which constructs the
memory by consolidating reasoning information over complementary temporal
ranges; and (2) \textbf{Memory Injection}, which makes the constructed memory
available to subsequent reasoning by feeding it back into selected decoder
layers.

\subsection{Preliminaries}

Let $\mathcal{M}$ denote a Multimodal Large Reasoning Model with $N$ decoder
layers. Given an image $I$ and a textual question $Q$, the model forms a
multimodal prompt $\mathcal{P}=(I,Q)$ and autoregressively generates a response
\begin{equation}
    \mathcal{Y}=(y_1,y_2,\ldots,y_T).
\end{equation}
The response contains a reasoning trajectory followed by a final answer. We
partition the reasoning trajectory into $K$ consecutive steps:
\begin{equation}
    \mathcal{R}
    =
    S_1 \oplus S_2 \oplus \cdots \oplus S_K,
    \qquad
    S_k=(y_{b_k},\ldots,y_{e_k}),
\end{equation}
where $\oplus$ denotes a boundary between consecutive reasoning steps, and
$b_k$ and $e_k$ denote the starting and ending positions of step $S_k$,
respectively. In our implementation, we segment the reasoning trajectory at
sentence terminators and newlines.

For token position $t$, let
$h_t^{(\ell)}\in\mathbb{R}^{d}$ denote the residual hidden state entering
decoder layer $\ell\in\{0,\ldots,N-1\}$. Omitting normalization operations for
clarity, the computation of layer $\ell$ is written as
\begin{align}
    a_t^{(\ell)}
    &=
    \operatorname{Attn}^{(\ell)}
    \!\left(
        h_{\leq t}^{(\ell)}
    \right), \\
    f_t^{(\ell)}
    &=
    \operatorname{FFN}^{(\ell)}
    \!\left(
        h_t^{(\ell)}+a_t^{(\ell)}
    \right), \\
    h_t^{(\ell+1)}
    &=
    h_t^{(\ell)}
    +
    a_t^{(\ell)}
    +
    f_t^{(\ell)}.
\end{align}
where $a_t^{(\ell)}$ and $f_t^{(\ell)}$ denote the attention and FFN outputs,
respectively. Our method uses $f_t^{(\ell)}$ from completed reasoning steps to
construct the memory and injects the resulting memory into
$h_t^{(\ell)}$ at subsequent step transitions.

\subsection{Recurrent Dual-Timescale Memory}
\label{sec:step-memory}

After reasoning step $S_k$ is completed, we extract a compact representation
from the FFN outputs near the end of the step using the tail window
$U_k=\{\max(b_k,e_k-w+1),\ldots,e_k\}$, which contains the last $w$ token
positions of $S_k$. At decoder layer $\ell$, the
step representation is obtained by mean pooling:
\begin{equation}
    s_k^{(\ell)}
    =
    \frac{1}{|U_k|}
    \sum_{t\in U_k}
    f_t^{(\ell)}.
    \label{eq:step_representation}
\end{equation}
The short tail window provides a more stable representation than a single
boundary token, while avoiding the mixture of intermediate FFN outputs
introduced by pooling the entire reasoning step. The resulting representation
provides a compact summary of the layer's transformed reasoning signal, rather
than copying the residual hidden state.

To preserve information from both recent and earlier reasoning steps, we
maintain an independent pair of recurrent memory states at each selected
decoder layer. Thus, memory states are not shared across layers; each layer
updates and later reads the memory constructed from its own step
representations:
\begin{align}
    m_{k,\mathrm{fast}}^{(\ell)}
    &=
    \lambda_{\mathrm{fast}}
    m_{k-1,\mathrm{fast}}^{(\ell)}
    +
    s_k^{(\ell)},
    \label{eq:fast_memory}\\
    m_{k,\mathrm{slow}}^{(\ell)}
    &=
    \lambda_{\mathrm{slow}}
    m_{k-1,\mathrm{slow}}^{(\ell)}
    +
    s_k^{(\ell)},
    \label{eq:slow_memory}
\end{align}
where $0\leq\lambda_{\mathrm{fast}}<\lambda_{\mathrm{slow}}<1$. Both memory
states are initialized to zero at the beginning of each input sequence. The
fast memory discounts earlier steps more aggressively and is therefore more
responsive to recently completed reasoning steps. In contrast, the slow memory
preserves contributions over a longer portion of the reasoning trajectory.

The two recurrent updates can be expanded as
\begin{equation}
    m_{k,q}^{(\ell)}
    =
    \sum_{j=1}^{k}
    \lambda_q^{k-j}s_j^{(\ell)},
    \qquad
    q\in\{\mathrm{fast},\mathrm{slow}\}.
    \label{eq:memory_expansion}
\end{equation}

Because $\lambda_{\mathrm{slow}}>\lambda_{\mathrm{fast}}$, the slow memory
retains historical updates more strongly and may grow larger in magnitude than
the fast memory. Before combining the two states, we constrain the norm of the
slow memory relative to the fast memory:
\begin{equation}
    \bar m_{k,\mathrm{slow}}^{(\ell)}
    =
    m_{k,\mathrm{slow}}^{(\ell)}
    \cdot
    \min
    \left(
        1,
        \frac{
            r
            \left\|
                m_{k,\mathrm{fast}}^{(\ell)}
            \right\|_2
        }{
            \left\|
                m_{k,\mathrm{slow}}^{(\ell)}
            \right\|_2
        }
    \right),
    \label{eq:slow_memory_cap}
\end{equation}
where $r>0$ specifies the maximum norm ratio. When the slow-memory norm is
below the threshold, the operation leaves it unchanged; otherwise, it rescales
only its magnitude while preserving its direction.

The final memory is constructed as
\begin{equation}
    m_k^{(\ell)}
    =
    m_{k,\mathrm{fast}}^{(\ell)}
    +
    \bar m_{k,\mathrm{slow}}^{(\ell)}.
    \label{eq:step_memory}
\end{equation}
The resulting memory balances responsiveness to recent reasoning with
longer-range retention over the preceding trajectory.

\subsection{Memory Injection}
\label{sec:memory-injection}
We inject the memory into decoder layers in the range
$\mathcal{L}_{\mathrm{inj}}=[L_{\min},L_{\max}]$. After step $S_k$ is completed, the
constructed memory is injected at the transition to the next reasoning step.
For $k<K$, we define the transition position as $\tau_k=e_k+1$, namely the
first decoding position following $S_k$. The causal boundary rule described in
the experimental protocol supplies $e_k$; no future tokens from $S_{k+1}$ are
needed to construct the memory for transition $\tau_k$.

For each selected layer $\ell\in\mathcal{L}_{\mathrm{inj}}$, we perform
additive memory injection:
\begin{equation}
    \widetilde h_{\tau_k}^{(\ell)}
    =
    h_{\tau_k}^{(\ell)}
    +
    \alpha m_k^{(\ell)},
    \label{eq:memory_injection}
\end{equation}
where $\alpha>0$ controls the injection strength.

The additive update may change the magnitude of the residual hidden state. We
therefore optionally rescale the injected state to preserve the original norm:
\begin{equation}
    \widehat h_{\tau_k}^{(\ell)}
    =
    \widetilde h_{\tau_k}^{(\ell)}
    \frac{
        \left\|
            h_{\tau_k}^{(\ell)}
        \right\|_2
    }{
        \left\|
            \widetilde h_{\tau_k}^{(\ell)}
        \right\|_2
    }.
    \label{eq:norm_preservation}
\end{equation}
The rescaled state is then used in place of the original residual hidden state.
Thus, norm preservation keeps the intervention approximately directional:
\methodname changes the activation direction while restoring its scale to that
of the unmodified state. For each transition, \methodname applies the following causal sequence: it pools
the last $w$ FFN outputs of the completed step, updates the fast and slow states,
caps the slow state if necessary, forms $m_k^{(\ell)}$, and injects it at
$\tau_k$ in the selected layers. The memory states are reset at the start of
each input and are carried across subsequent reasoning steps within that input.

\begin{table*}[!t]
    \centering
    \setlength{\tabcolsep}{1mm}
    \renewcommand{\arraystretch}{1.04}
    {\small
    \begin{tabular*}{\textwidth}{@{\extracolsep{\fill}}lcccc@{\hspace{12pt}}cccc@{\hspace{12pt}}ccc@{}}
        \toprule
        \multirow{2}{*}{Method}
        & \multicolumn{4}{c}{Mathematics}
        & \multicolumn{4}{c}{Science}
        & \multicolumn{3}{c}{General VQA} \\
        \cmidrule(lr){2-5}
        \cmidrule(lr){6-9}
        \cmidrule(lr){10-12}
        & MathVision & MathVerse & MMK12-Math & Avg.
        & Physics & Chemistry & Biology & Avg.
        & MMStar & MMMU-val & Avg. \\
        \midrule
        \multicolumn{12}{c}{\textit{Qwen3-VL-4B}} \\
        Vanilla & 42.42 & 58.79 & 75.67 & 58.96 & 70.00 & 67.40 & 67.73 & 68.38 & \underline{69.17} & 62.98 & \underline{66.08} \\
        MemVR  & 43.58 & 56.73 & \underline{76.33} & 58.88 & 69.80 & 64.80 & 68.20 & 67.60 & 68.88 & 63.18 & 66.03 \\
        VISTA   & 43.20  & 58.41 & 75.07 & 58.89 & \underline{70.33} & 67.60  & \textbf{69.47}  & \underline{69.13} & 68.65 & 61.00 & 64.83 \\
        ECRD    & \underline{44.07} & 57.82 & 75.00 & 58.96 & \underline{70.33} & 67.07 & 67.73 & 68.38 & 69.02 & 62.92 & 65.97 \\
        LEAD    & 43.59 & \underline{59.50} & 75.80 & \underline{59.63} & 69.20 & \underline{67.67} & 68.93 & 68.60 & 68.33 & \underline{63.22} & 65.78 \\
        \cdashline{1-12}
        \methodname & \textbf{44.96} & \textbf{59.69} & \textbf{76.60} & \textbf{60.42} & \textbf{70.60} & \textbf{68.33} & \underline{69.20} & \textbf{69.38} & \textbf{69.96} & \textbf{63.33} & \textbf{66.65} \\
        \addlinespace
        \midrule
        \multicolumn{12}{c}{\textit{Qwen3-VL-8B}} \\
        Vanilla & 47.58 & 59.60 & 77.00 & 61.39 & \underline{72.13} & 71.60 & 74.00 & 72.58 & 70.60 & 65.83 & 68.22 \\
        MemVR  & 46.05 & 59.97 & 77.60 & 61.21 & 71.60 & 72.40 & 74.30 & 72.77 & 70.20 & 66.10 & 68.15 \\
        VISTA   & \underline{49.34} & \underline{60.02} & 77.46 & 62.27 & 71.26 & 72.40 & 74.26 & 72.64 & \underline{70.91} & \textbf{66.98} & \textbf{68.95} \\
        ECRD    & 48.68 & 59.98 & \underline{78.20} & \underline{62.29} & 72.07 & \underline{73.33} & \underline{75.53} & \underline{73.64} & 70.78 & 66.09 & 68.44 \\
        LEAD    & 48.85 & 59.95 & 76.80 & 61.87 & 72.10 & 73.07 & 75.00 & 73.39 & 70.70 & 65.83 & 68.27 \\
        \cdashline{1-12}
        \methodname & \textbf{49.67} & \textbf{60.57} & \textbf{78.40} & \textbf{62.88} & \textbf{73.46} & \textbf{73.67} & \textbf{75.60} & \textbf{74.24} & \textbf{71.09} & \underline{66.38} & \underline{68.74} \\
        \addlinespace
        \midrule
        \multicolumn{12}{c}{\textit{InternVL3.5-4B}} \\
        Vanilla & 43.31 & \underline{52.15} & 67.80 & 54.42 & 61.60 & 75.60 & 65.40 & 67.53 & 68.07 & 64.79 & 66.43 \\
        MemVR  & 44.72 & 50.13 & \underline{68.80} & \underline{54.55} & \underline{62.60} & 75.07 & 66.40 & 68.02 & 68.73 & 65.50 & 67.12 \\
        VISTA   & 43.53 & 51.26 & 67.06 & 53.95 & 61.67 & \underline{75.73} & 67.13 & 68.18 & 69.04 & 65.21 & 67.13 \\
        ECRD    & 44.40 & 49.20 & 67.53 & 53.71 & 59.53 & 75.07 & 64.27 & 66.29 & \underline{69.24} & \underline{65.65} & \underline{67.45} \\
        LEAD    & \underline{44.73} & 48.48 & 68.27 & 53.83 & 62.27 & 75.06 & \underline{67.80} & \underline{68.38} & 68.41 & 65.47 & 66.94 \\
        \cdashline{1-12}
        \methodname & \textbf{45.39} & \textbf{52.41} & \textbf{70.20} & \textbf{56.00} & \textbf{63.20} & \textbf{76.53} & \textbf{68.07} & \textbf{69.27} & \textbf{69.27} & \textbf{65.81} & \textbf{67.54} \\
        \addlinespace
        \midrule
        \multicolumn{12}{c}{\textit{InternVL3.5-8B}} \\
        Vanilla & 47.47 & 52.40 & \underline{77.40} & \underline{59.09} & 65.00 & 78.80 & 71.46 & 71.75 & 70.94 & 68.47 & 69.71 \\
        MemVR  & 45.28 & 52.48 & 76.60 & 58.12 & 64.80 & 78.60 & \textbf{72.60} & 72.00 & 70.88 & 68.82 & 69.85 \\
        VISTA   & 45.94 & 51.81 & 77.13 & 58.29 & \textbf{65.53} & 78.87 & \underline{72.33} & \underline{72.24} & 70.08 & 68.82 & 69.45 \\
        ECRD    & 47.27 & 52.41 & \textbf{77.60} & \underline{59.09} & 64.13 & \textbf{79.20} & 71.53 & 71.62 & 70.82 & 68.19 & 69.51 \\
        LEAD    & \underline{47.69} & \underline{52.70} & 76.60 & 59.00 & 64.73 & \textbf{79.20} & 72.20 & 72.04 & \underline{71.08} & \textbf{69.14} & \underline{70.11} \\
        \cdashline{1-12}
        \methodname & \textbf{48.02} & \textbf{52.79} & \textbf{77.60} & \textbf{59.47} & \underline{65.20} & \underline{79.00} & \textbf{72.60} & \textbf{72.27} & \textbf{71.43} & \underline{68.95} & \textbf{70.19} \\
        \bottomrule
    \end{tabular*}
    }
    \caption{Main results on mathematical, scientific, and general visual
    reasoning benchmarks. Bold and underlined values indicate the best and
    second-best results within each backbone.}
    \label{tab:specific_results}
\end{table*}

\section{Experiments}

\subsection{Experimental Setup}

\textbf{Benchmarks and Models.} We evaluate \methodname on eight benchmarks
grouped into three categories: Mathematical Reasoning (MathVision
\citep{wang2024measuring}, MathVerse \citep{zhang2024mathverse}, and
MMK12-Math), Scientific Reasoning (MMK12-Phys, MMK12-Chem, and MMK12-Bio)
\citep{meng2025mm}, and General Visual Question
Answering (MMStar \citep{chen2024we} and MMMU \citep{yue2024mmmu}). For hallucination evaluation, we
additionally use MMVP \citep{tong2024eyes} and MME-Hall \citep{fu2026mme}. We use
Qwen3-VL-4B/8B \citep{bai2025qwen3} and InternVL3.5-4B/8B
\citep{wang2025internvl3}, and report \textbf{Accuracy} with
higher values indicating better performance. Further details are provided in the Appendix.

\textbf{Baselines.}
We compare \methodname with unmodified \textbf{Vanilla Decoding} and four representative
training-free test-time methods that strengthen visual evidence during
generation. \textbf{VISTA} steers generation using activation signals and early-layer
visual logits \citep{li2025hidden}. \textbf{MemVR} reinjects visual features
into intermediate representations at uncertainty-triggered layers
\citep{zou2025look}. \textbf{LEAD} combines uncertainty-aware switching between
latent and discrete reasoning with visual-anchor reinjection
\citep{xu2026thinking}, while \textbf{ECRD} iteratively guides decoding with an
image-derived evidence pool \citep{zhang2026see}.

\textbf{Implementation Details.}
We segment reasoning steps at sentence terminators and newlines. We run all
models in vLLM~\citep{kwon2023efficient}, cap generation at 4,096 tokens, use a temperature of $0.6$,
and report accuracy averaged over three runs. Additional implementation details are provided in the Appendix.

\subsection{Main Results}
\label{sec:results-analysis}

Table~\ref{tab:specific_results} summarizes the main results across four
backbones and eight visual reasoning benchmarks. Compared with vanilla
decoding, \methodname improves accuracy on all individual backbone--benchmark
settings, covering mathematical, scientific, and general visual reasoning
tasks. The improvement is especially consistent on the two 4B backbones, where
\methodname obtains the best result on most benchmarks and improves all three
task-group averages. These results show a broad mean-accuracy benefit across
model families and task types. Compared with existing training-free intervention
methods, \methodname also achieves strong overall performance. On Qwen3-VL-4B
and InternVL3.5-4B, \methodname obtains the best average result in all three
benchmark groups. On Qwen3-VL-8B, it achieves the best mathematical and
scientific averages and the second-best General VQA average. The gains are
particularly clear on mathematical and
scientific reasoning, where solving the task often requires maintaining
intermediate relations, constraints, and conclusions over multiple steps. This
is aligned with the design of \methodname, which constructs and reuses memory
from the generated reasoning trajectory. On InternVL3.5-8B, \methodname remains competitive with the strongest baseline.
Although LEAD obtains slightly higher scores on Chemistry and MMMU, \methodname
performs better on the other benchmarks and achieves the best grouped averages.
Overall, the results indicate that recurrent trajectory memory is an effective
training-free strategy for improving test-time multimodal reasoning.

\subsection{Hallucination Evaluation}
\label{sec:hallucination}

Since \methodname intervenes in hidden representations during decoding, we
further evaluate whether the accuracy improvements come at the cost of visual
faithfulness. We report results on MMVP and MME-Hall in
Table~\ref{tab:hallucination}. Overall, \methodname preserves hallucination
performance across backbones while improving visual reasoning accuracy in the
main evaluation. On the two InternVL3.5 backbones, \methodname improves both
MME-Hall and MMVP scores. On the two Qwen3-VL backbones, the results remain
largely stable, with only small changes relative to vanilla decoding. These
results indicate that injecting trajectory-derived memory does not introduce a
systematic degradation in visual faithfulness.

\begin{table}[t]
    \centering
    \setlength{\tabcolsep}{1mm}
    \renewcommand{\arraystretch}{1.08}
    {\small
    \begin{tabular}{lcccccc}
        \toprule
        \textbf{Method} & \multicolumn{5}{c}{\textbf{MME-Hall}} & \textbf{MMVP} \\
        \cmidrule(lr){2-6}
        & \textbf{Exist.} & \textbf{Count} & \textbf{Pos.} & \textbf{Color} & \textbf{Total} & \\
        \midrule
        \multicolumn{7}{c}{\textit{Qwen3-VL-4B}} \\
        Vanilla & \textbf{193.33} & \textbf{174.44} & \textbf{161.67} & \textbf{187.22} & \textbf{716.67} & \textbf{82.11} \\
        \methodname & \textbf{193.33} & \textbf{174.44} & 160.00 & \textbf{187.22} & 715.00 & 81.78 \\
        \midrule
        \multicolumn{7}{c}{\textit{Qwen3-VL-8B}} \\
        Vanilla & \textbf{193.33} & \textbf{170.56} & 161.11 & \textbf{193.33} & 718.33 & \textbf{79.89} \\
        \methodname & \textbf{193.33} & \textbf{170.56} & \textbf{161.67} & \textbf{193.33} & \textbf{718.89} & 79.67 \\
        \midrule
        \multicolumn{7}{c}{\textit{InternVL3.5-4B}} \\
        Vanilla & \textbf{190.00} & 173.33 & 154.44 & \textbf{176.67} & 694.44 & 74.44 \\
        \methodname & \textbf{190.00} & \textbf{175.00} & \textbf{156.11} & \textbf{176.67} & \textbf{697.78} & \textbf{74.67} \\
        \midrule
        \multicolumn{7}{c}{\textit{InternVL3.5-8B}} \\
        Vanilla & \textbf{195.00} & 169.44 & \textbf{170.00} & 186.67 & 721.11 & 78.78 \\
        \methodname & \textbf{195.00} & \textbf{170.56} & \textbf{170.00} & \textbf{191.67} & \textbf{727.22} & \textbf{81.00} \\
        \bottomrule
    \end{tabular}
    }
    \caption{Hallucination results on MME-Hall and MMVP; ``Total'' sums the four
    MME-Hall subtasks.}
    \label{tab:hallucination}
\end{table}

\subsection{Ablation Study}
\label{sec:ablation}


\begin{figure*}[!htbp]
    \centering
    \includegraphics[width=0.99\textwidth]{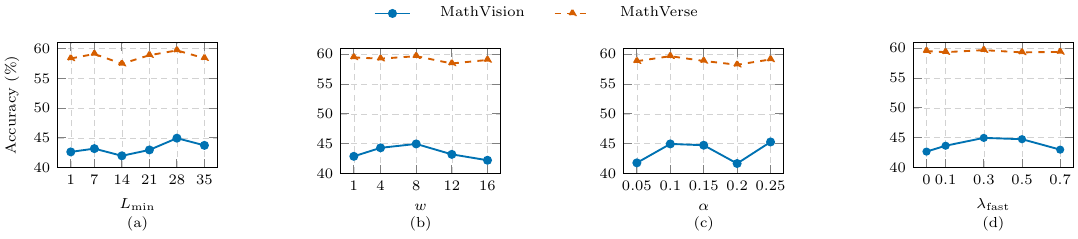}
    \caption{Hyperparameter sensitivity of \methodname on MathVision and
    MathVerse. The four panels vary the injection layer, window size,
    intervention strength, and fast-decay factor, respectively.}
    \label{fig:ablation}
\end{figure*}

\subsubsection{Component Ablation}
\label{sec:component-ablation}

We evaluate each component by removing the fast branch, the slow branch, and
the two stabilization operations from the default configuration. The slow-memory
norm cap corresponds to Eq.~\eqref{eq:slow_memory_cap}, and output-norm
preservation corresponds to Eq.~\eqref{eq:norm_preservation}. As shown in
Table~\ref{tab:ablation_components}, the full \methodname achieves
the best results on both benchmarks. Removing either the fast or slow branch
leads to a clear drop on MathVision, showing that both short-term and longer-term
trajectory information are useful. The stabilization operations also contribute,
especially output-norm preservation on MathVision. On MathVerse, the changes are
smaller, but the full configuration still performs best.

\begin{table}[!htbp]
    \centering
    \setlength{\tabcolsep}{4pt}
    \renewcommand{\arraystretch}{1.08}
    \begin{tabular}{lcc}
        \toprule
        \textbf{Method} & MathVision & MathVerse \\
        \midrule
        \textsc{Vanilla} & 42.42 & 58.79 \\
        \midrule
        \methodname & \textbf{44.96} & \textbf{59.69} \\
        w/o fast & 42.65 & 59.52 \\
        w/o slow & 42.87 & 59.09 \\
        w/o slow norm cap & 44.84 & 59.35 \\
        w/o norm preservation & 43.20 & 59.48 \\
        \bottomrule
    \end{tabular}
    \caption{Component ablation on MathVision and MathVerse using
    Qwen3-VL-4B.}
    \label{tab:ablation_components}
\end{table}

\subsubsection{Effect of Injection Layer}
\label{sec:ablation-injection-layer}

Figure~\ref{fig:ablation}(a) shows that starting memory injection from upper
decoder layers gives better results than intervening from earlier layers. The
best setting is $L_{\min}=28$, while starting too early or too late leads to
lower accuracy. This suggests that memory injection is more effective when
applied to higher-level representations while still leaving several layers for
subsequent integration.

\subsubsection{Effect of Window Size}
\label{sec:ablation-window}

Figure~\ref{fig:ablation}(b) compares different window sizes for reading local
hidden states around each detected boundary. A moderate window size, $w=8$,
performs best on both benchmarks. Smaller windows may provide insufficient
context for summarizing a reasoning step, while larger windows do not bring
additional gains.

\subsubsection{Effect of Intervention Strength}
\label{sec:ablation-strength}

Figure~\ref{fig:ablation}(c) evaluates the intervention strength $\alpha$.
A moderate strength works better than very weak or overly strong intervention.
When $\alpha$ is too small, the memory signal may not sufficiently affect
subsequent decoding; when it is too large, the injected memory can overly
perturb the model's original hidden representations. This suggests that
\methodname benefits from a controlled memory injection rather than aggressive
state modification.

\subsubsection{Effect of the Decay Factor}
\label{sec:ablation-fast-decay}

Figure~\ref{fig:ablation}(d) studies the fast decay factor
$\lambda_{\mathrm{fast}}$ with the slow decay fixed. A moderate value,
$\lambda_{\mathrm{fast}}=0.3$, gives the best overall result. When the value is
too small, the fast branch carries limited recent information; when it is too
large, the fast state becomes less distinct from the slower memory stream. The
point at zero corresponds to the \textit{w/o fast} ablation.

\subsection{Further Analysis}
\label{sec:further-analysis}

\subsubsection{When Is the Memory Most Useful?}
\label{sec:length-analysis}

\begin{figure}[!htbp]
    \centering
    \includegraphics[width=0.98\linewidth]{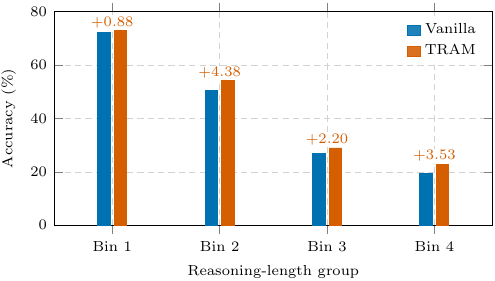}
    \caption{Performance across reasoning lengths on MathVision using
    Qwen3-VL-4B. The four groups are obtained by sorting paired traces
    by their mean reasoning length, from shortest to longest.}
    \label{fig:length_analysis}
\end{figure}

We further analyze how the effect of \methodname changes with reasoning length.
For each question, we pair the vanilla output and the output generated with
\methodname, and assign the pair to one of four bins according to their average
trace length. Figure~\ref{fig:length_analysis} shows that \methodname improves
accuracy in all length groups. The gain is modest in the shortest group
(+0.88), while the three longer groups show larger improvements (+4.38, +2.20,
and +3.53). This suggests that trajectory-derived memory becomes more useful
when the reasoning trace is longer and contains more intermediate information
to preserve.


\subsubsection{How Should the Memory Be Read and Constructed?}
\label{sec:memory-design-analysis}

We further ablate how the memory state is read and constructed in
Table~\ref{tab:memory_design}. For read location, using either the attention
output or the residual output leads to lower performance than the default
setting, showing that the choice of where to extract the state matters for
constructing useful memory. For memory construction, visual summaries, fixed chunks, single-step variants,
and random noise all perform worse than the full design. Fixed chunks use a
similar average update frequency as the default detector but place boundaries
periodically, which does not match the structure of the generated reasoning.
The previous-step-only and current-step-only variants also lag behind the full
model, indicating that recurrent aggregation across multiple reasoning steps is
important. Overall, these results show that \methodname benefits from both
trajectory-aware boundary detection and recurrent memory construction.

\begin{table}[!htbp]
    \centering
    \setlength{\tabcolsep}{2.2pt}
    \renewcommand{\arraystretch}{1.06}
    \begin{tabular}{llcc}
        \toprule
        \textbf{Group} & \textbf{Variant} &
        MathVision & MathVerse \\
        \midrule
        \textbf{Baseline} & Vanilla & 42.42 & 58.79 \\
        & \methodname (default) & \textbf{44.96} & \textbf{59.69} \\
        \midrule
        \multirow{2}{*}{Read location}
            & Attn. output & 40.24 & 59.05 \\
            & Residual output & 41.22 & 54.70 \\
        \midrule
        \multirow{5}{*}{Construction}
            & Visual summary & 43.31 & \underline{59.09} \\
            & Fixed chunks & 42.10 & 58.80 \\
            & Previous step only & 43.31 & 58.71 \\
            & Current step only & \underline{43.97} & 58.97 \\
            & Random noise & 41.99 & 58.62 \\
        \bottomrule
    \end{tabular}
    \caption{Memory design ablations on MathVision and MathVerse using
    Qwen3-VL-4B.}
    \label{tab:memory_design}
\end{table}

\subsubsection{Inference Latency}
\label{sec:latency}

We measure decoding efficiency on MathVision using a single NVIDIA RTX A6000
48 GB GPU, with GPU memory utilization set to 0.95. Vanilla decoding and
\methodname use the same models, batch sizes, attention backends, prompts, and
generation budgets. Table~\ref{tab:latency} reports the latency and peak
allocated GPU memory for all four backbones. \methodname introduces a small
latency increase of 0.87 to 1.98 ms per generated token, while peak memory
remains unchanged at the reported precision. This is because \methodname keeps
only compact recurrent states rather than storing trajectory-level hidden
states.

\begin{table}[!htbp]
    \centering
    \setlength{\tabcolsep}{2.4pt}
    \renewcommand{\arraystretch}{1.05}
    {\normalsize
    \begin{tabular}{lcccc}
        \toprule
        \multirow{2}{*}{Backbone} &
        \multicolumn{2}{c}{Latency (ms/token)} &
        \multicolumn{2}{c}{Peak memory (GB)} \\
        \cmidrule(lr){2-3} \cmidrule(lr){4-5}
        & Vanilla & \methodname & Vanilla & \methodname \\
        \midrule
        Qwen3-VL-4B & 23.47 & 25.45 & 44.66 & 44.66 \\
        Qwen3-VL-8B & 28.10 & 29.11 & 44.46 & 44.46 \\
        InternVL3.5-4B & 18.72 & 20.06 & 44.73 & 44.73 \\
        InternVL3.5-8B & 27.27 & 28.14 & 44.63 & 44.63 \\
        \bottomrule
    \end{tabular}
    }
    \caption{Decode latency and peak allocated GPU memory on MathVision using a
    single NVIDIA RTX A6000 48 GB GPU.}
    \label{tab:latency}
\end{table}

\section{Related Work}

\paragraph{Multimodal Reasoning.}
Recent reinforcement learning with verifiable rewards (RLVR) has strengthened
long-form multimodal reasoning on tasks requiring visual understanding and
multi-step inference
\citep{guo2025deepseek,huang2025vision,meng2025mm}.
Some methods incorporate explicit visual operations, such as zooming, cropping,
drawing, or region selection, into the reasoning process
\citep{hu2024visual,su2025openthinkimg}.
Others optimize visual grounding during RL by reinforcing visually relevant
tokens or trajectories
\citep{wang2026visually,huang2025spotlight}. Inference-time methods further
strengthen grounding through visually conditioned decoding calibration and
visual-representation steering
\citep{leng2024mitigating,li2025hidden,xu2026thinking}.
Other methods revisit the input image during reasoning
\citep{ghosal2026visref,gao2025interleaved,zhang2026see}.
By contrast, \methodname consolidates completed reasoning into an auxiliary
memory to retain information formed throughout the trajectory.
%

\paragraph{Memory-Augmented Reasoning.}
Long-context reasoning requires models to retain and integrate information
across multiple steps, while standard Transformers lack a dedicated mechanism
for managing such information beyond the growing token context
\citep{omidi2025memory}.
Drawing inspiration from human memory, some methods augment Transformer
architectures with dedicated memory modules to support multi-step inference
and integrate information distributed over long contexts
\citep{hwang2024transformerfam,cheng2026conditional}.
In multimodal reasoning, several methods maintain visual representations as
memory through learned retrieval pathways, allowing visual evidence to be
reused across reasoning steps
\citep{huang2026persistent}. VisMem further learns short- and long-term latent
visual memories through dedicated memory construction and invocation modules
\citep{yu2025vismem}. Training-free methods instead reuse visual evidence
through inference-time reinjection or iterative evidence extraction
\citep{zou2025look,zhang2026see}.
This line of work has largely centered on preserving access to visual evidence
throughout reasoning. Motivated by our attribution analysis, we introduce
\methodname, a training-free trajectory-derived auxiliary memory that
consolidates completed reasoning states during generation for reuse in
subsequent reasoning.

\section{Conclusion}

We presented \textbf{\methodname}, a training-free method that augments
standard decoding with a compact auxiliary memory constructed from the model's
own reasoning trajectory. \methodname consolidates completed reasoning into a
compact latent memory, updates it through fast and slow recurrent streams, and
feeds it into selected decoder layers through a lightweight residual pathway.
Experiments across four MLRM variants on eight benchmarks spanning
mathematical, scientific, and general visual reasoning show that \methodname
improves mean accuracy over vanilla decoding without additional training or
parameter updates. Together with our attribution analysis, these results
support retaining and reusing information formed during completed reasoning as
an explicit resource for long-form multimodal reasoning.


\bibliography{aaai2027}

@misc{bai2025qwen3,
      title={Qwen3-VL Technical Report}, 
      author={Shuai Bai and Yuxuan Cai and Ruizhe Chen and Keqin Chen and Xionghui Chen and Zesen Cheng and Lianghao Deng and Wei Ding and Chang Gao and Chunjiang Ge and Wenbin Ge and Zhifang Guo and Qidong Huang and Jie Huang and Fei Huang and Binyuan Hui and Shutong Jiang and Zhaohai Li and Mingsheng Li and Mei Li and Kaixin Li and Zicheng Lin and Junyang Lin and Xuejing Liu and Jiawei Liu and Chenglong Liu and Yang Liu and Dayiheng Liu and Shixuan Liu and Dunjie Lu and Ruilin Luo and Chenxu Lv and Rui Men and Lingchen Meng and Xuancheng Ren and Xingzhang Ren and Sibo Song and Yuchong Sun and Jun Tang and Jianhong Tu and Jianqiang Wan and Peng Wang and Pengfei Wang and Qiuyue Wang and Yuxuan Wang and Tianbao Xie and Yiheng Xu and Haiyang Xu and Jin Xu and Zhibo Yang and Mingkun Yang and Jianxin Yang and An Yang and Bowen Yu and Fei Zhang and Hang Zhang and Xi Zhang and Bo Zheng and Humen Zhong and Jingren Zhou and Fan Zhou and Jing Zhou and Yuanzhi Zhu and Ke Zhu},
      year={2025},
      eprint={2511.21631},
      archivePrefix={arXiv},
      primaryClass={cs.CV},
      url={https://arxiv.org/abs/2511.21631}, 
}

@misc{wang2025internvl3,
      title={InternVL3.5: Advancing Open-Source Multimodal Models in Versatility, Reasoning, and Efficiency}, 
      author={Weiyun Wang and Zhangwei Gao and Lixin Gu and Hengjun Pu and Long Cui and Xingguang Wei and Zhaoyang Liu and Linglin Jing and Shenglong Ye and Jie Shao and Zhaokai Wang and Zhe Chen and Hongjie Zhang and Ganlin Yang and Haomin Wang and Qi Wei and Jinhui Yin and Wenhao Li and Erfei Cui and Guanzhou Chen and Zichen Ding and Changyao Tian and Zhenyu Wu and Jingjing Xie and Zehao Li and Bowen Yang and Yuchen Duan and Xuehui Wang and Zhi Hou and Haoran Hao and Tianyi Zhang and Songze Li and Xiangyu Zhao and Haodong Duan and Nianchen Deng and Bin Fu and Yinan He and Yi Wang and Conghui He and Botian Shi and Junjun He and Yingtong Xiong and Han Lv and Lijun Wu and Wenqi Shao and Kaipeng Zhang and Huipeng Deng and Biqing Qi and Jiaye Ge and Qipeng Guo and Wenwei Zhang and Songyang Zhang and Maosong Cao and Junyao Lin and Kexian Tang and Jianfei Gao and Haian Huang and Yuzhe Gu and Chengqi Lyu and Huanze Tang and Rui Wang and Haijun Lv and Wanli Ouyang and Limin Wang and Min Dou and Xizhou Zhu and Tong Lu and Dahua Lin and Jifeng Dai and Weijie Su and Bowen Zhou and Kai Chen and Yu Qiao and Wenhai Wang and Gen Luo},
      year={2025},
      eprint={2508.18265},
      archivePrefix={arXiv},
      primaryClass={cs.CV},
      url={https://arxiv.org/abs/2508.18265}, 
}

@inproceedings{
huang2025vision,
title={Vision-R1: Incentivizing Reasoning Capability in Multimodal Large Language Models},
author={Wenxuan Huang and Bohan Jia and Shaosheng Cao and Zheyu Ye and Fei zhao and Zhe Xu and Yao Hu and Shaohui Lin},
booktitle={The Fourteenth International Conference on Learning Representations},
year={2026},
url={https://openreview.net/forum?id=UZIjskfbfU}
}

@article{
meng2025mm,
title={{MM}-Eureka: Toward Stable Multimodal Reasoning via Rule-based Reinforcement Learning with Policy Drift Control},
author={Fanqing Meng and Lingxiao Du and Zongkai Liu and Zhixiang Zhou and Quanfeng Lu and Tiancheng Han and Daocheng Fu and Kaipeng Zhang and Ping Luo and Yu Qiao and Jiaheng Zhang and Michael Qizhe Shieh and Qiaosheng Zhang and Wenqi Shao},
journal={Transactions on Machine Learning Research},
issn={2835-8856},
year={2026},
url={https://openreview.net/forum?id=8y1ch6y24H},
note={}
}

@article{liu2024lost,
  title={Lost in the middle: How language models use long contexts},
  author={Liu, Nelson F and Lin, Kevin and Hewitt, John and Paranjape, Ashwin and Bevilacqua, Michele and Petroni, Fabio and Liang, Percy},
  journal={Transactions of the association for computational linguistics},
  volume={12},
  pages={157--173},
  year={2024}
}

@inproceedings{
liu2026sinktrack,
title={SinkTrack: Attention Sink based Context Anchoring for Large Language Models},
author={Xu Liu and Guikun Chen and Wenguan Wang},
booktitle={The Fourteenth International Conference on Learning Representations},
year={2026},
url={https://openreview.net/forum?id=Gg1aPETCL6}
}

@inproceedings{wang2026visually,
  title={Visually-guided policy optimization for multimodal reasoning},
  author={Wang, Zengbin and Xiong, Feng and Lin, Liang and Hu, Xuecai and Wang, Yong and Wang, Yanlin and Zhang, Man and Chu, Xiangxiang},
  booktitle={Proceedings of the 64th Annual Meeting of the Association for Computational Linguistics (Volume 1: Long Papers)},
  pages={6646--6664},
  year={2026}
}

@misc{huang2026persistent,
      title={Persistent Visual Memory: Sustaining Perception for Deep Generation in LVLMs}, 
      author={Siyuan Huang and Xiaoye Qu and Yafu Li and Tong Zhu and Zefeng He and Muxin Fu and Daizong Liu and Wei-Long Zheng and Yu Cheng},
      year={2026},
      eprint={2605.00814},
      archivePrefix={arXiv},
      primaryClass={cs.CV},
      url={https://arxiv.org/abs/2605.00814}, 
}

@inproceedings{ghosal2026visref,
  title={VisRef: Visual Refocusing while Thinking Improves Test-Time Scaling in Multi-Modal Large Reasoning Models},
  author={Ghosal, Soumya Suvra and Kim, Youngeun and Li, Zhuowei and Chaudhry, Ritwick and Xu, Linghan and Zhang, Hongjing and Zablocki, Jakub and Xing, Yifan and Zhang, Qin},
  booktitle={Proceedings of the IEEE/CVF Conference on Computer Vision and Pattern Recognition},
  pages={33404--33414},
  year={2026}
}

@inproceedings{yang2026look,
  title={Look-back: Implicit visual re-focusing in mllm reasoning},
  author={Yang, Shuo and Niu, Yuwei and Liu, Yuyang and Ye, Yang and Lin, Bin and Yuan, Li},
  booktitle={Proceedings of the AAAI conference on artificial intelligence},
  volume={40},
  pages={11694--11702},
  year={2026}
}

@inproceedings{
huang2025spotlight,
title={Spotlight on Token Perception for Multimodal Reinforcement Learning},
author={Siyuan Huang and Xiaoye Qu and Yafu Li and Yun Luo and Zefeng He and Daizong Liu and Yu Cheng},
booktitle={The Fourteenth International Conference on Learning Representations},
year={2026},
url={https://openreview.net/forum?id=bRA4lVWJVQ}
}

@inproceedings{gao2025interleaved,
  title={Interleaved-modal chain-of-thought},
  author={Gao, Jun and Li, Yongqi and Cao, Ziqiang and Li, Wenjie},
  booktitle={Proceedings of the Computer Vision and Pattern Recognition Conference},
  pages={19520--19529},
  year={2025}
}

@inproceedings{leng2024mitigating,
  title={Mitigating object hallucinations in large vision-language models through visual contrastive decoding},
  author={Leng, Sicong and Zhang, Hang and Chen, Guanzheng and Li, Xin and Lu, Shijian and Miao, Chunyan and Bing, Lidong},
  booktitle={Proceedings of the IEEE/CVF Conference on Computer Vision and Pattern Recognition},
  pages={13872--13882},
  year={2024}
}

@inproceedings{
li2025hidden,
title={The Hidden Life of Tokens: Reducing Hallucination of Large Vision-Language Models Via Visual Information Steering},
author={Zhuowei Li and Haizhou Shi and Yunhe Gao and Di Liu and Zhenting Wang and Yuxiao Chen and Ting Liu and Long Zhao and Hao Wang and Dimitris N. Metaxas},
booktitle={Forty-second International Conference on Machine Learning},
year={2025},
url={https://openreview.net/forum?id=7BKcLeHQsm}
}

@inproceedings{zou2025look,
  title={Look Twice Before You Answer: Memory-Space Visual Retracing for Hallucination Mitigation in Multimodal Large Language Models},
  author={Zou, Xin and Wang, Yizhou and Yan, Yibo and Lyu, Yuanhuiyi and Zheng, Kening and Huang, Sirui and Chen, Junkai and Jiang, Peijie and Liu, Jia and Tang, Chang and others},
  booktitle={International Conference on Machine Learning},
  pages={80873--80899},
  year={2025},
  organization={PMLR}
}

@InProceedings{zhang2026see,
    author    = {Zhang, Yongchang and Ma, Oliver and Liu, Tianyi and Zhou, Guangquan and Chen, Yang},
    title     = {See It, Say It, Sorted: An Iterative Training-Free Framework for Visually-Grounded Multimodal Reasoning in LVLMs},
    booktitle = {Proceedings of the IEEE/CVF Conference on Computer Vision and Pattern Recognition (CVPR)},
    month     = {June},
    year      = {2026},
    pages     = {11933-11942}
}

@article{guo2025deepseek,
  title={DeepSeek-R1 incentivizes reasoning in LLMs through reinforcement learning},
  author={Guo, Daya and Yang, Dejian and Zhang, Haowei and Song, Junxiao and Wang, Peiyi and Zhu, Qihao and Xu, Runxin and Zhang, Ruoyu and Ma, Shirong and Bi, Xiao and others},
  journal={Nature},
  volume={645},
  number={8081},
  pages={633--638},
  year={2025},
  publisher={Nature Publishing Group UK London}
}

@inproceedings{hu2024visual,
 author = {Hu, Yushi and Shi, Weijia and Fu, Xingyu and Roth, Dan and Ostendorf, Mari and Zettlemoyer, Luke and Smith, Noah and Krishna, Ranjay},
 booktitle = {Advances in Neural Information Processing Systems},
 doi = {10.52202/079017-4423},
 editor = {A. Globerson and L. Mackey and D. Belgrave and A. Fan and U. Paquet and J. Tomczak and C. Zhang},
 pages = {139348--139379},
 publisher = {Curran Associates, Inc.},
 title = {Visual Sketchpad: Sketching as a Visual Chain of Thought for Multimodal Language Models},
 url = {https://proceedings.neurips.cc/paper_files/paper/2024/file/fb82011040977c7712409fbdb5456647-Paper-Conference.pdf},
 volume = {37},
 year = {2024}
}

@misc{su2025openthinkimg,
      title={OpenThinkIMG: Learning to Think with Images via Visual Tool Reinforcement Learning}, 
      author={Zhaochen Su and Linjie Li and Mingyang Song and Yunzhuo Hao and Zhengyuan Yang and Jun Zhang and Guanjie Chen and Jiawei Gu and Juntao Li and Xiaoye Qu and Yu Cheng},
      year={2025},
      eprint={2505.08617},
      archivePrefix={arXiv},
      primaryClass={cs.CV},
      url={https://arxiv.org/abs/2505.08617}, 
}

@misc{pan2026towards,
      title={Towards Long-Horizon Interpretability: Efficient and Faithful Multi-Token Attribution for Reasoning LLMs}, 
      author={Wenbo Pan and Zhichao Liu and Xianlong Wang and Haining Yu and Xiaohua Jia},
      year={2026},
      eprint={2602.01914},
      archivePrefix={arXiv},
      primaryClass={cs.LG},
      url={https://arxiv.org/abs/2602.01914}, 
}

@InProceedings{xu2026thinking,
    author    = {Xu, Zhongxing and Wang, Zhonghua and Qian, Zhe and Shi, Dachuan and Tang, Feilong and Hu, Ming and Su, Shiyan and Zou, Xiaocheng and Feng, Wei and Mahapatra, Dwarikanath and Peng, Yifan and Lin, Minquan and Ge, Zongyuan},
    title     = {Thinking in Uncertainty: Mitigating Hallucinations in MLRMs with Latent Entropy-Aware Decoding},
    booktitle = {Proceedings of the IEEE/CVF Conference on Computer Vision and Pattern Recognition (CVPR)},
    month     = {June},
    year      = {2026},
    pages     = {11064-11075}
}

@InProceedings{yu2025vismem,
    author    = {Yu, Xinlei and Xu, Chengming and Zhang, Guibin and Chen, Zhangquan and Zhang, Yudong and He, Yongbo and Jiang, Peng-Tao and Zhang, Jiangning and Hu, Xiaobin and Yan, Shuicheng},
    title     = {VisMem: Latent Vision Memory Unlocks Potential of Vision-Language Models},
    booktitle = {Proceedings of the IEEE/CVF Conference on Computer Vision and Pattern Recognition (CVPR)},
    month     = {June},
    year      = {2026},
    pages     = {31544-31555}
}

@misc{omidi2025memory,
      title={Memory-Augmented Transformers: A Systematic Review from Neuroscience Principles to Enhanced Model Architectures}, 
      author={Parsa Omidi and Xingshuai Huang and Axel Laborieux and Bahareh Nikpour and Tianyu Shi and Armaghan Eshaghi},
      year={2025},
      eprint={2508.10824},
      archivePrefix={arXiv},
      primaryClass={cs.LG},
      url={https://arxiv.org/abs/2508.10824}, 
}

@misc{hwang2024transformerfam,
      title={TransformerFAM: Feedback attention is working memory}, 
      author={Dongseong Hwang and Weiran Wang and Zhuoyuan Huo and Khe Chai Sim and Pedro Moreno Mengibar},
      year={2024},
      eprint={2404.09173},
      archivePrefix={arXiv},
      primaryClass={cs.LG},
      url={https://arxiv.org/abs/2404.09173}, 
}

@inproceedings{cheng2026conditional,
  title={Conditional memory via scalable lookup: A new axis of sparsity for large language models},
  author={Cheng, Xin and Zeng, Wangding and Dai, Damai and Chen, Qinyu and Wang, Bingxuan and Xie, Zhenda and Huang, Kezhao and Yu, Xingkai and Hao, Zhewen and Zhang, Han and others},
  booktitle={Proceedings of the 64th Annual Meeting of the Association for Computational Linguistics (Volume 1: Long Papers)},
  pages={4968--4990},
  year={2026}
}

@article{wang2024measuring,
  title={Measuring multimodal mathematical reasoning with math-vision dataset},
  author={Wang, Ke and Pan, Junting and Shi, Weikang and Lu, Zimu and Ren, Houxing and Zhou, Aojun and Zhan, Mingjie and Li, Hongsheng},
  journal={Advances in Neural Information Processing Systems},
  volume={37},
  pages={95095--95169},
  year={2024}
}

@inproceedings{jain2019attention,
  title={Attention is not explanation},
  author={Jain, Sarthak and Wallace, Byron C},
  booktitle={Proceedings of the 2019 Conference of the North American Chapter of the Association for Computational Linguistics: Human Language Technologies, Volume 1 (Long and Short Papers)},
  pages={3543--3556},
  year={2019}
}

@inproceedings{zhang2024mathverse,
  title={Mathverse: Does your multi-modal llm truly see the diagrams in visual math problems?},
  author={Zhang, Renrui and Jiang, Dongzhi and Zhang, Yichi and Lin, Haokun and Guo, Ziyu and Qiu, Pengshuo and Zhou, Aojun and Lu, Pan and Chang, Kai-Wei and Qiao, Yu and others},
  booktitle={European Conference on Computer Vision},
  pages={169--186},
  year={2024},
  organization={Springer}
}

@article{chen2024we,
  title={Are we on the right way for evaluating large vision-language models?},
  author={Chen, Lin and Li, Jinsong and Dong, Xiaoyi and Zhang, Pan and Zang, Yuhang and Chen, Zehui and Duan, Haodong and Wang, Jiaqi and Qiao, Yu and Lin, Dahua and others},
  journal={Advances in Neural Information Processing Systems},
  volume={37},
  pages={27056--27087},
  year={2024}
}

@inproceedings{yue2024mmmu,
  title={Mmmu: A massive multi-discipline multimodal understanding and reasoning benchmark for expert agi},
  author={Yue, Xiang and Ni, Yuansheng and Zhang, Kai and Zheng, Tianyu and Liu, Ruoqi and Zhang, Ge and Stevens, Samuel and Jiang, Dongfu and Ren, Weiming and Sun, Yuxuan and others},
  booktitle={Proceedings of the IEEE/CVF conference on computer vision and pattern recognition},
  pages={9556--9567},
  year={2024}
}

@inproceedings{tong2024eyes,
  title={Eyes wide shut? exploring the visual shortcomings of multimodal llms},
  author={Tong, Shengbang and Liu, Zhuang and Zhai, Yuexiang and Ma, Yi and LeCun, Yann and Xie, Saining},
  booktitle={Proceedings of the IEEE/CVF conference on computer vision and pattern recognition},
  pages={9568--9578},
  year={2024}
}

@article{fu2026mme,
  title={Mme: A comprehensive evaluation benchmark for multimodal large language models},
  author={Fu, Chaoyou and Chen, Peixian and Shen, Yunhang and Qin, Yulei and Zhang, Mengdan and Lin, Xu and Yang, Jinrui and Zheng, Xiawu and Li, Ke and Sun, Xing and others},
  journal={Advances in Neural Information Processing Systems},
  volume={38},
  year={2026}
}

@inproceedings{kwon2023efficient,
  title={Efficient memory management for large language model serving with pagedattention},
  author={Kwon, Woosuk and Li, Zhuohan and Zhuang, Siyuan and Sheng, Ying and Zheng, Lianmin and Yu, Cody Hao and Gonzalez, Joseph and Zhang, Hao and Stoica, Ion},
  booktitle={Proceedings of the 29th symposium on operating systems principles},
  pages={611--626},
  year={2023}
}

\end{document}